\documentclass[conference]{IEEEtran}
\IEEEoverridecommandlockouts
\usepackage{cite}
\usepackage{amsmath,amssymb,amsfonts}
\usepackage{algorithmic}
\usepackage{graphicx}
\usepackage{textcomp}
\usepackage[dvipsnames]{xcolor}
\usepackage{url}
\usepackage{soul}
\usepackage{xspace}
\usepackage{multirow,tabularx}
\usepackage{subfigure}
\usepackage{arydshln}
\usepackage{booktabs}
\usepackage{xcolor}
\def\BibTeX{{\rm B\kern-.05em{\sc i\kern-.025em b}\kern-.08em
    T\kern-.1667em\lower.7ex\hbox{E}\kern-.125emX}}
\usepackage{multirow}

\begin{document}

\title{Decompose, Enrich, and Extract! Schema-aware Event Extraction using LLMs.}

\author{\IEEEauthorblockN{1\textsuperscript{nd} Fatemeh Shiri}
\IEEEauthorblockA{\textit{Department of Data Science and AI} \\
\textit{Faculty of Information Technology}\\
Monash University, Australia \\
Fatemeh.Shiri@monash.edu}
\and
\IEEEauthorblockN{2\textsuperscript{th} Van Nguyen}
\IEEEauthorblockA{\textit{Information Sciences Division} \\
\textit{Defence Science and Technology Group}\\
Australia \\
Van.Nguyen5@defence.gov.au}
\and
\IEEEauthorblockN{3\textsuperscript{rd} Farhad Moghimifar}
\IEEEauthorblockA{\textit{Department of Data Science and AI} \\
\textit{Faculty of Information Technology}\\
Monash University, Australia \\
Farhad.Moghimifar@monash.edu}
\and
\IEEEauthorblockN{4\textsuperscript{th} John Yoo}
\IEEEauthorblockA{\textit{Information Sciences Division} \\
\textit{Defence Science and Technology Group}\\
Australia \\
John.Yoo@defence.gov.au}
\and
\IEEEauthorblockN{5\textsuperscript{rd} Gholamreza Haffari}
\IEEEauthorblockA{\textit{Department of Data Science and AI} \\
\textit{Faculty of Information Technology}\\
Monash University, Australia \\
Gholamreza.Haffari@monash.edu}
\and
\IEEEauthorblockN{6\textsuperscript{rd} Yuan-Fang Li}
\IEEEauthorblockA{\textit{Department of Data Science and AI} \\
\textit{Faculty of Information Technology}\\
Monash University, Australia \\
YuanFang.Li@monash.edu}
}


\maketitle

\begin{abstract}
Large Language Models (LLMs) demonstrate significant capabilities in processing natural language data, promising efficient knowledge extraction from diverse textual sources to enhance situational awareness and support decision-making. However, concerns arise due to their susceptibility to hallucination, resulting in contextually inaccurate content. This work focuses on harnessing LLMs for automated Event Extraction, introducing a new method to address hallucination by decomposing the task into Event Detection and Event Argument Extraction. Moreover, the proposed method integrates dynamic schema-aware augmented retrieval examples into prompts tailored for each specific inquiry, thereby extending and adapting advanced prompting techniques such as Retrieval-Augmented Generation. Evaluation findings on prominent event extraction benchmarks and results from a synthesized benchmark illustrate the method's superior performance compared to baseline approaches. 
\end{abstract}


\section{Introduction}
Enhancing strategic awareness for effective military decision-making requires processing and analysing information from various data sources. One category of data sources of particular interest in this work is textual data available in vast quantities in the public domain, such as online news articles and reports. However, handling such data sources poses significant challenges, including issues of volume and relevance \cite{shiri2021toward}. Manual processing of the large volume and complexity of open-source data leads to cognitive overload. However, automated extraction of knowledge using traditional Natural Language Processing (NLP) techniques is also a challenging task.

Recent advances in Artificial Intelligence, especially with the emergence of Large Language Models (LLMs), have demonstrated an impressive ability to process and comprehend natural language at a sophisticated level\cite{achiam2023gpt}. This capability enables them to execute tasks like language generation, translation, summarisation, and beyond. Consequently, they offer promising potential in efficiently processing and analysing substantial volumes of data, crucial for augmenting human decision-making and military planning \cite{jensen2023large}. Specifically, LLMs can be harnessed to automatically extract structured knowledge, such as in the form of event extraction (EE). This not only aids human analysts in visualising and describing complex situations in the form of a knowledge graph, but also facilitates automated downstream reasoning and learning for generating actionable insights. 

Despite numerous remarkable breakthroughs, the hallucination problem associated with LLMs raises significant concerns, particularly in adopting these technologies in the defense context\cite{ji2023survey,li2023halueval}. Hallucination refers to generating content that strays from factual reality or includes fabricated information.
This work is focused on leveraging LLMs for automating event extraction, toward supporting the creation of knowledge graphs and enhancing decision-making processes. To enhance the relevance and accuracy of the outcomes, we proposes a new method to mitigate the hallucination problem. This is achieved by adapting state-of-the-art prompting approaches, such as Chain-of-Thought~\cite{cot} and Retrieval Augmented Generation~\cite{lewis2020retrieval}.

\begin{figure}[t]
    \centering
    \resizebox{.4\textwidth}{!}{
    \includegraphics{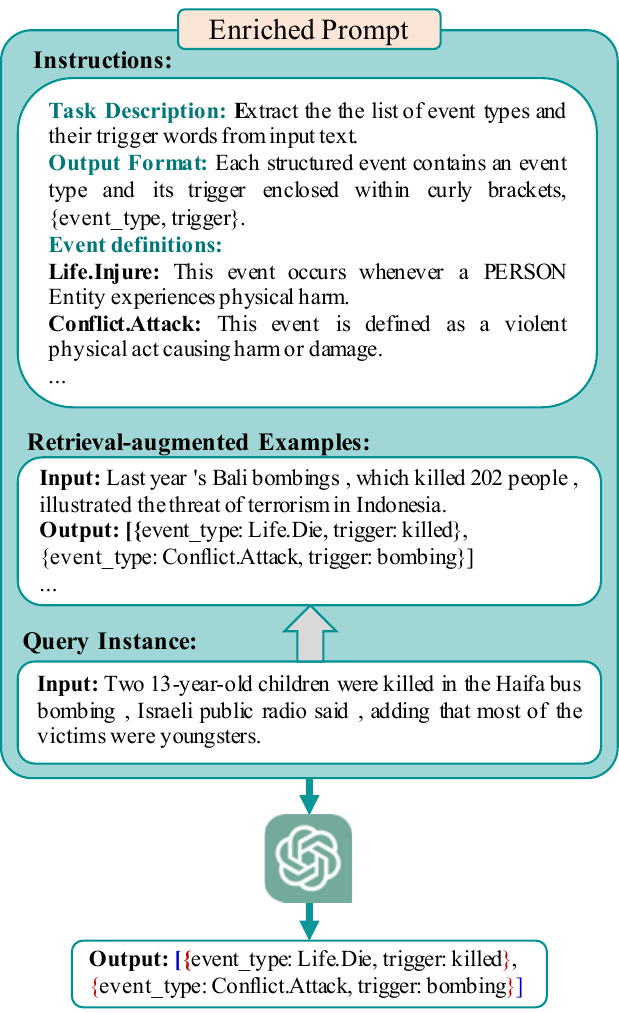}}
    \caption{An example of enriched prompt for event extraction using GPT-4. GPT-4 is tasked with query instances based on provided instructions, event type definitions, output format, and retrieval-augmented examples in this scenario. These examples are the most similar instances to the query instance retrieved from the existing knowledge base. GPT-4 is expected to produce responses for each query instance without any prior training on the specific task or data. (For simplicity, we only show the first step, Event Detection).}
    \label{fig:open}
\end{figure}

EE is a challenging NLP task that requires the identification and extraction of structured event records from unstructured text. It typically includes a trigger indicating the occurrence of the event and multiple arguments of pre-defined roles. As mentioned earlier, EE plays a crucial role as it provides valuable information for various downstream tasks, including knowledge graph construction \cite{zhang2020aser} and question answering \cite{han2021ester}.

State-of-the-art generative event extraction approaches rely on fine-tuning sequence-to-sequence models \cite{text2event,moghimifar2023theia}. These models require large-scale annotated data which is expensive and time-consuming to obtain. The emergence of LLMs such as ChatGPT and GPT-4 \cite{achiam2023gpt} provides an opportunity to solve language tasks using in-context learning (ICL) without the need for task-specific datasets and fine-tuning \cite{brown2020language}. ICL leverages the concept of ``learning by demonstration,'' a method where the pre-trained model is able to recognize and replicate tasks by giving it a few examples. This phenomenon allows the model to mimic task-specific behavior by following the instructions and adjusting its generated responses to match the demonstrated examples.
By providing a well-devised prompt, LLMs can therefore learn to perform EE tasks. However, the challenge remains as to how to devise an appropriate input prompt to be able to extract complex structured events while minimizing the risk of hallucination. 

To mitigate hallucinations generated by LLMs, we decompose the EE task into two steps: Event Detection (ED) and Event Argument Extraction (EAE). We utilize schema-aware prompts, including granular instructions and the most relevant demonstration examples, called retrieval-augmented examples, tailored to the query instance.

Figure \ref{fig:open} displays an example of event detection performed by the GPT-4 model using a retrieval-augmented prompt. The input instance and the target events are both taken from the ACE2005 dataset \cite{ace2005}.

In summary, our contributions are as follows:
\begin{itemize}
\item We construct automatic retrieval-augmented prompts for event detection and argument extraction sub-tasks on both high-resource and low-resource situations.
\item  Instead of relying only on task description and output options, our enriched prompt integrates extraction rules, specifies output formats, and provides retrieval-augmented examples to provide LLMs with an in-depth understanding of the structured extraction tasks.
\item We synthesize a new event benchmark called MaritimeEvent with 10,000 data points. We evaluate our proposed framework on popular EE benchmarks and our MaritimeEvent benchmark. The results demonstrate its advantages over baselines. 
\item Through detailed analysis and case study, we demonstrate the essential role of retrieval augmentation in boosting EE performance.
\end{itemize}

\section{Background}
Our approach is related to two lines of research.

\subsection{Generative Event Extraction}
Traditional event extraction methods required fine-grained (token-level and entity-level) annotations \cite{oneie}, \cite{nguyen2019one}, \cite{shiri2019recovering}. Recent generative approaches tackle this problem as a sequence-to-sequence learning problem referred to as End-to-End event extraction \cite{text2event,alta}.
TANL \cite{TANL} and GRIT \cite{GRIT} employ neural generation models for event extraction by focusing on sequence generation. While TEXT2EVENT \cite{text2event} and KC-GEE \cite{KC-GEE} focus on structure generation, TEXT2EVENT directly generates event schema and text spans to form event records via constrained decoding.

\subsection{Event Extraction with LLMs}
LLMs are central to NLP due to their impressive performance on numerous tasks \cite{wei2022emergent,mahowald2023dissociating,moghimifar2023few}. LLMs are inherently context-sensitive, they may produce different responses based on the specific formulation of the prompt. By carefully crafting and refining prompts,  the capabilities of LLM can be fully utilized, tailoring the outputs to match the nuances of diverse tasks and objectives.

Chain-of-Thought (CoT) \cite{cot} is an enhanced prompting approach designed to enhance the proficiency of LLMs in handling intricate reasoning tasks, including arithmetic reasoning \cite{cobbe2021training}, commonsense reasoning  \cite{cot}.

Despite the potential benefits of LLMs in context learning (ICL), existing literature has limitations in evaluating these models’ efficacy for this task. The most relevant to our research is leveraging ChatGPT for the information
extraction task \cite{li2023evaluating,han2023information}, including event detection \cite{sharif2023characterizing}, and event argument extraction \cite{zhang2024ultra,wei2023zero}. These papers mainly focus on either curating new benchmark datasets \cite{gao2023benchmarking} or benchmarking ChatGPT’s performance with simple instructions without detailed extraction guidelines. The latter reports results that are inferior to those achieved by specialized supervised EE systems\cite{li2023evaluating,han2023information}. 

Instead of using a single instruction to extract multiple information types, \cite{gao2023benchmarking} propose fine-grained IE prompting for each information type extraction. For example, in the case of entity extraction, they manually craft fine-grained instructions for each event type and argument in the dataset. However, this approach demands substantial effort and results in lengthy context prompts for LLMs. In contrast, we propose a prompting strategy that includes granular instructions along with automatic retrieval augmented examples which are vital for adapting to EE tasks.

\section{Approach}\label{sec:approach}
\begin{figure*}[t]
    \centering
    \resizebox{.9\textwidth}{!}{
    \includegraphics{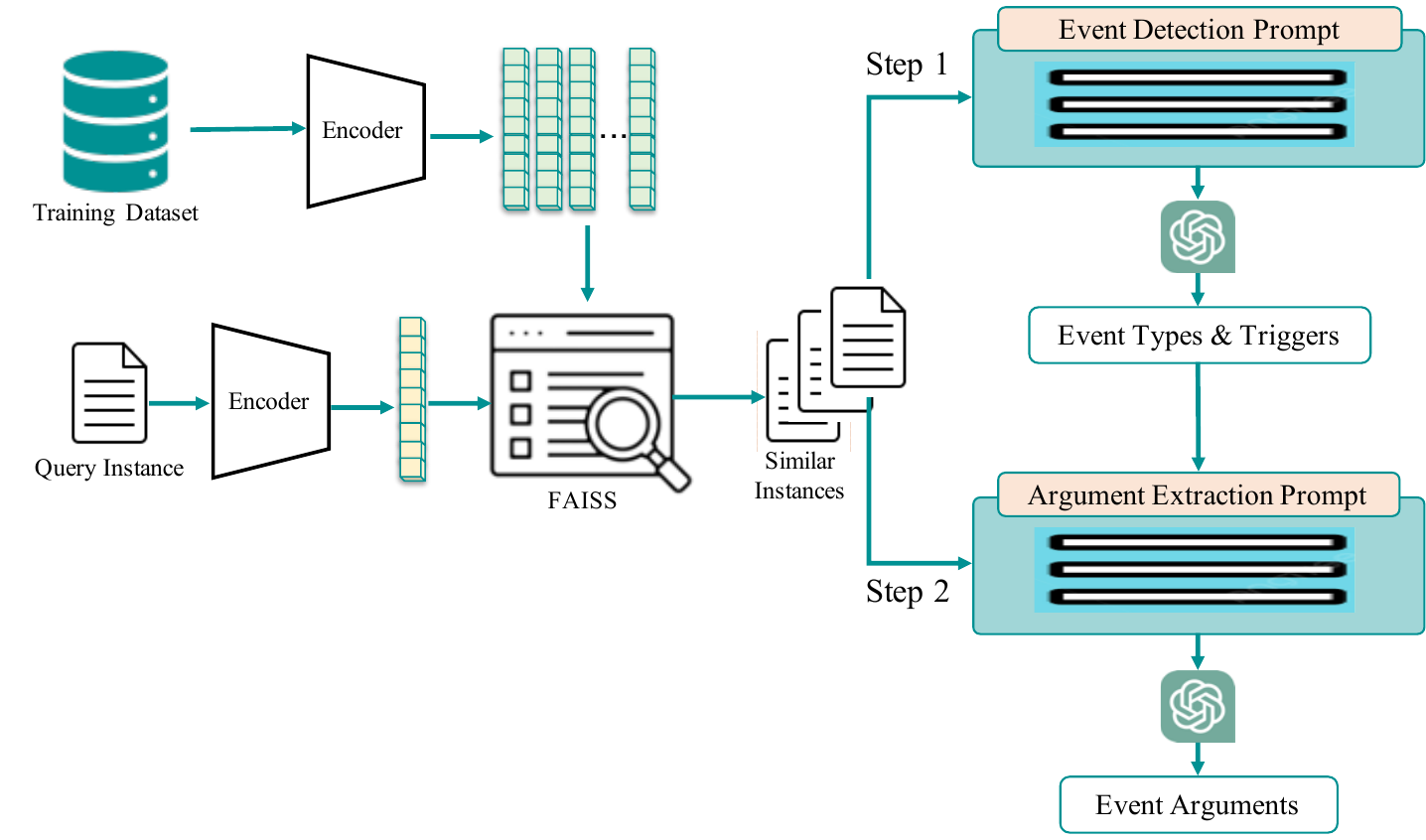}}
    \caption{An illustration of our end-to-end framework for event extraction, which performs event detection and event argument extraction jointly. The details of the Event Detection Prompt and Argument Extraction Prompt are shown in Figures \ref{fig:ed_prompt} and \ref{fig:eae_prompt}.}
    \label{fig:approach}
\end{figure*}

We propose an end-to-end framework that extracts the structured events records from input text in two steps using LLMs. Figure \ref{fig:approach} illustrates a high-level overview of our approach. 
Given the query instance, our system generates the embedding representation of the query instance and uses Facebook AI Similarity Search (FAISS) \cite{johnson2019billion} to prioritize and retrieve the top $K$ instances from the training dataset that demonstrate the highest similarity to the query. These retrieved examples, combined with the granular instructions in both the ED prompt and EAE prompt, are fed into the LLMs for event extraction. In the following, we explain each component of our framework in detail.

\subsection{Task Decomposition}
LLMs have advanced text input capabilities, yet encounter challenges in retrieving accurate information from lengthy contexts \cite{junqing2023never, xu2023retrieval}. The ``lost in the middle'' issue significantly impacts their accuracy, especially when accurate information is positioned within the middle of the text. Extracting structured events from text, especially from documents, necessitates providing task instructions, schema, extraction rules, event and argument definitions, output format, and demonstration examples. However, this lengthy prompt may challenge LLMs, potentially leading to inaccuracies in event extraction.

Decomposed prompting was proposed to solve complex tasks by decomposing tasks (via prompting) into simpler sub-tasks \cite{khot2022decomposed}. Each sub-task is tackled in sequence, with the outcome appended before progressing to the next \cite{drozdov2022compositional}. 
Inspired by this research direction, we address the challenge of long-context comprehension in LLMs by breaking down Event Extraction (EE) into two sub-tasks: ED which focuses on identifying event triggers and event types, and EAE which deals with extracting event arguments for predefined roles. For each sub-task, we design specific prompts, along with demonstration examples. Figures \ref{fig:ed_prompt} and \ref{fig:eae_prompt} showcase two examples of our enriched prompts for the ED and EAE sub-tasks, respectively. We examine the efficacy of this decomposed prompting approach within the context of EE in Section \ref{sec:decompose}.

\subsection{Granular Instructions}
An essential function of LLMs is their ability to execute natural language instructions, often referred to as zero-shot prompts \cite{wei2021finetuned}. The capability of LLMs to precisely understand and execute natural language instructions, especially structured information extraction is vital, ensuring both the accuracy of the outcomes and the reliability of their executions. Ambiguous or inadequate instructions can result in unintended outcomes, potentially leading to serious consequences. Therefore, it is crucial to offer granular and comprehensive instructions when interacting with an LLM.

\begin{figure*}[t]
    \centering
    \resizebox{.9\textwidth}{!}{
    \includegraphics{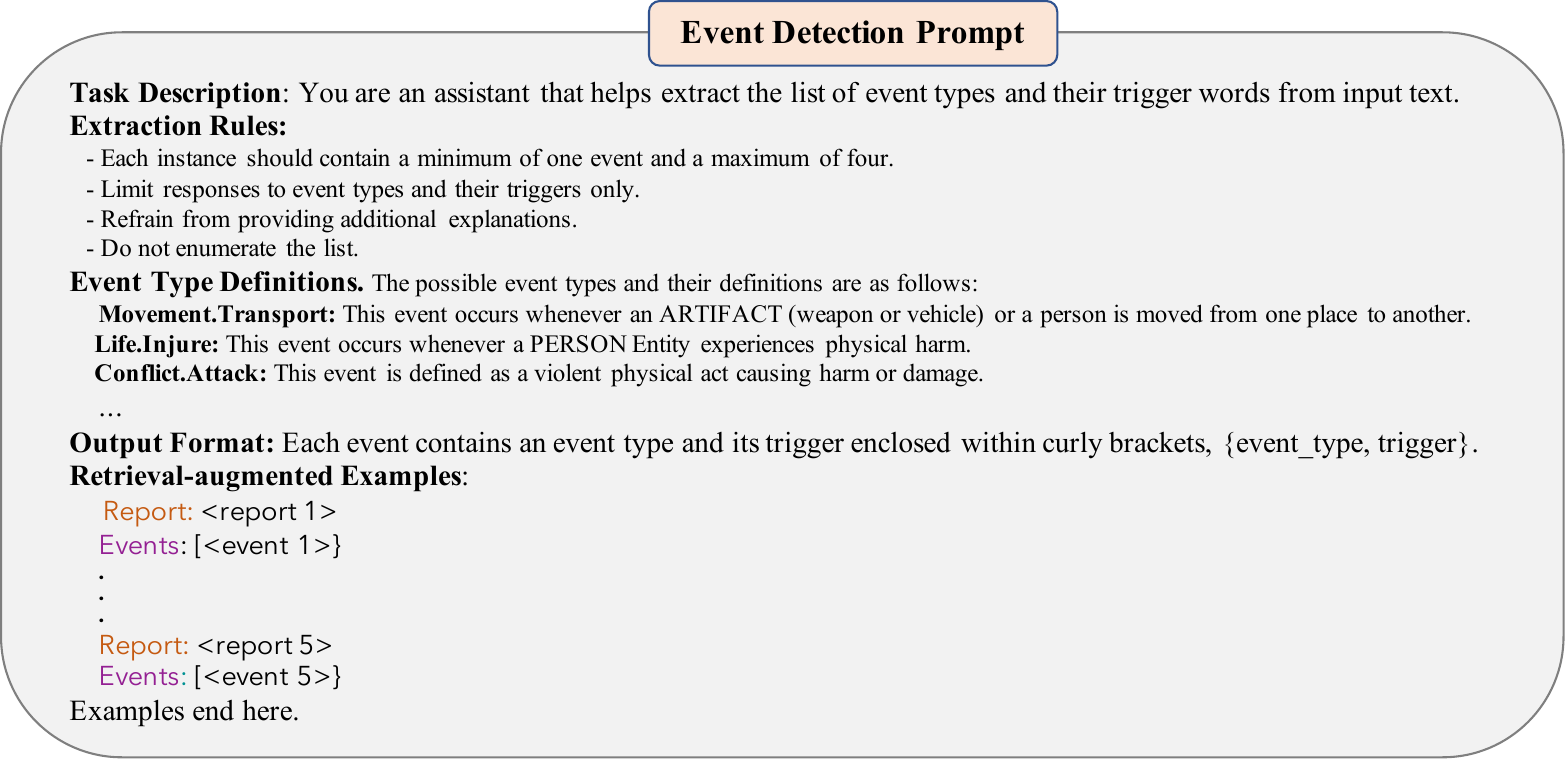}}
    \caption{An illustration of our granular Instructions and placeholder for retrieval augmented examples within the ED prompt}
    \label{fig:ed_prompt}
\end{figure*}

\begin{figure*}[t]
    \centering
    \resizebox{.9\textwidth}{!}{
    \includegraphics{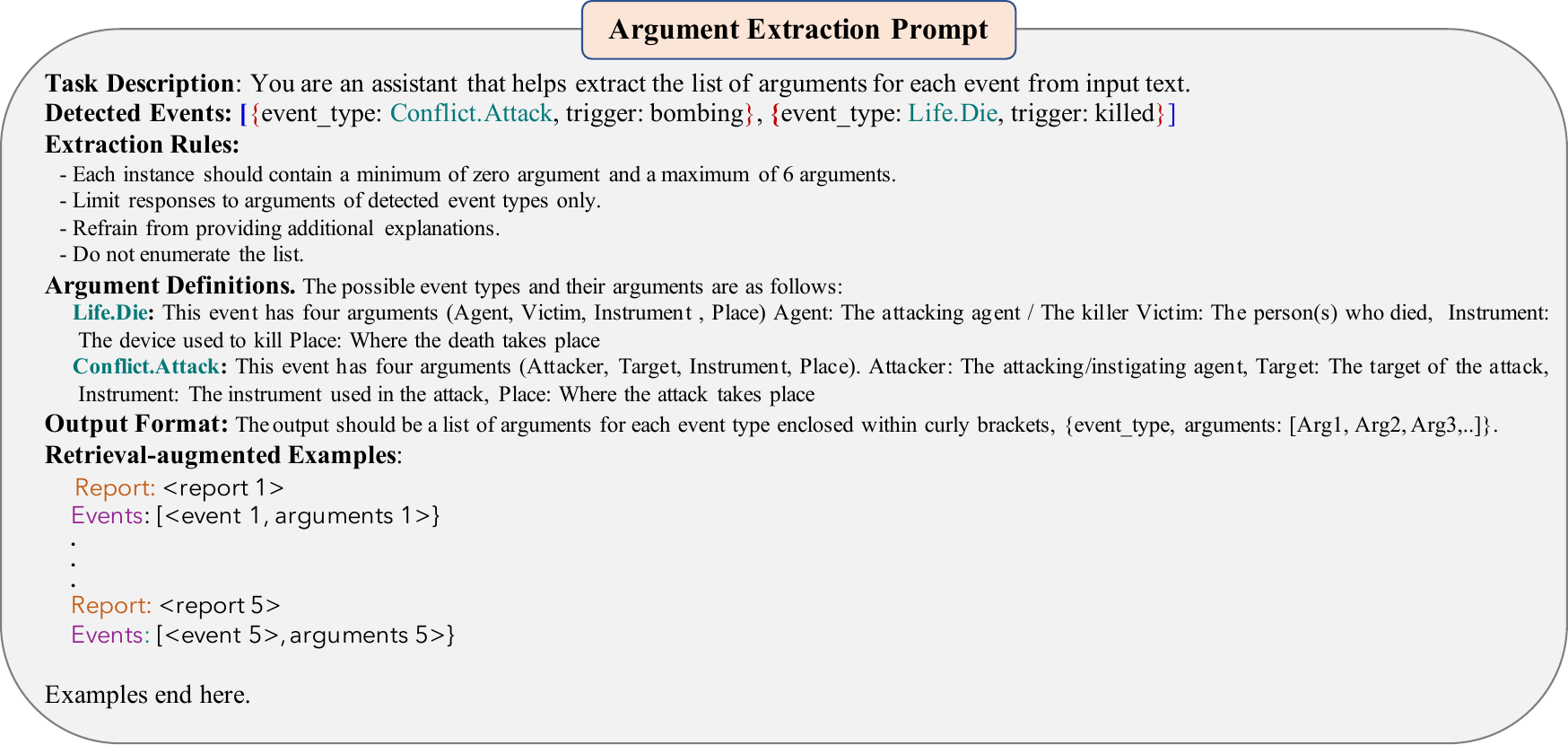}}
    \caption{An illustration of our granular Instructions and placeholder for retrieval augmented examples within the EAE prompt}
    \label{fig:eae_prompt}
\end{figure*}

In order to extract accurate structured events from text, we provide enriched prompts including task description, schema, extraction rules, event types and argument definitions, output format, and demonstration examples for each sub-task. Figures \ref{fig:ed_prompt} and \ref{fig:eae_prompt} show detailed instructions of both ED and EAE sub-tasks.

\subsection{Retrieval Augmented Examples (RAE) for Prompt Enriching}

 
A significant benefit of ICL, in contrast to traditional fine-tuning methods, is its decreased dependency on extensive amounts of annotated data. A straightforward yet potent implementation of ICL is through one-shot and few-shot prompting. These methods employ standardized examples that remain consistent across all input instances, irrespective of their relevance. However, this may lead to a lack of flexibility in understanding diverse contexts and specific input variations, resulting in suboptimal performance. 

We address this problem by dynamically adapting and refining LLMs' understanding based on the specific context of each input instance. While ICL does not necessitate a large scale labeled dataset as in fine-tuning methods, the inclusion of a retrieval mechanism allows such external data, if available, to serve as a valuable resource. This facilitates the discovery of contextually relevant examples for input instances, thereby enhancing the performance of ICL. The retrieval of examples from the existing labeled data for the EE task can be facilitated through several search mechanisms. The choice of a specific example retriever is non-trivial, as it significantly impacts the retrieved examples and their subsequent contribution to task execution. Drawing inspiration from Retrieval Augmented Generation (RAG) \cite{lewis2020retrieval}, which incorporates the retrieval of external data into the generative process, we retrieve instances from the existing set corresponding to each query. Our approach mirrors the conventional process of indexing, retrieval, and generation, aligning with the methodology employed in RAG.

The first step in our approach is to transform existing instances into vector representations through an embedding model. We employ three different embedding techniques discussed in section \ref{sec:eval}. Next, the encoded instance vectors are indexed using the ${IndexFlatL2}$ structure provided by FAISS.\footnote{FAISS is an open-source library for dense vector clustering and similarity search.} Upon receipt of a user query, the system employs the same encoding model utilized during the indexing phase to transform the input query into a vector representation. It then proceeds to compute the similarity scores between the query vector and the vectorized existing instances within the indexed corpus. The system prioritizes and retrieves the top $K$ instances that demonstrate the highest similarity to the query. Finally, the retrieved context combined with the prompt is fed into the LLMs for event extraction.


\section{Experiments}

\subsection{Experimental Setup}
\textbf{Dataset.} 
We carry out experiments on two popular event extraction datasets: the sentence-level dataset Automatic Content Extraction 2005 (ACE05-EN) \cite{ace2005}, and the document-level dataset: \textsc{WikiEvents} \cite{wikievent}. 
It is worth noting that \textsc{WikiEvents} presents significant challenges due to three factors. (1) Each instance in ACE05-EN contains only one sentence, whereas instances in \textsc{WikiEvents} are documents. (2) Almost every instance in ACE05-EN contains only one event, whereas multiple events could be present in one instance in \textsc{WikiEvents}. (3) The amount of training data in ACE05-EN is more than 77 times greater than that in \textsc{WikiEvents}. 

\begin{figure*}[t]
    \centering
    \resizebox{0.85\textwidth}{!}{
    \includegraphics{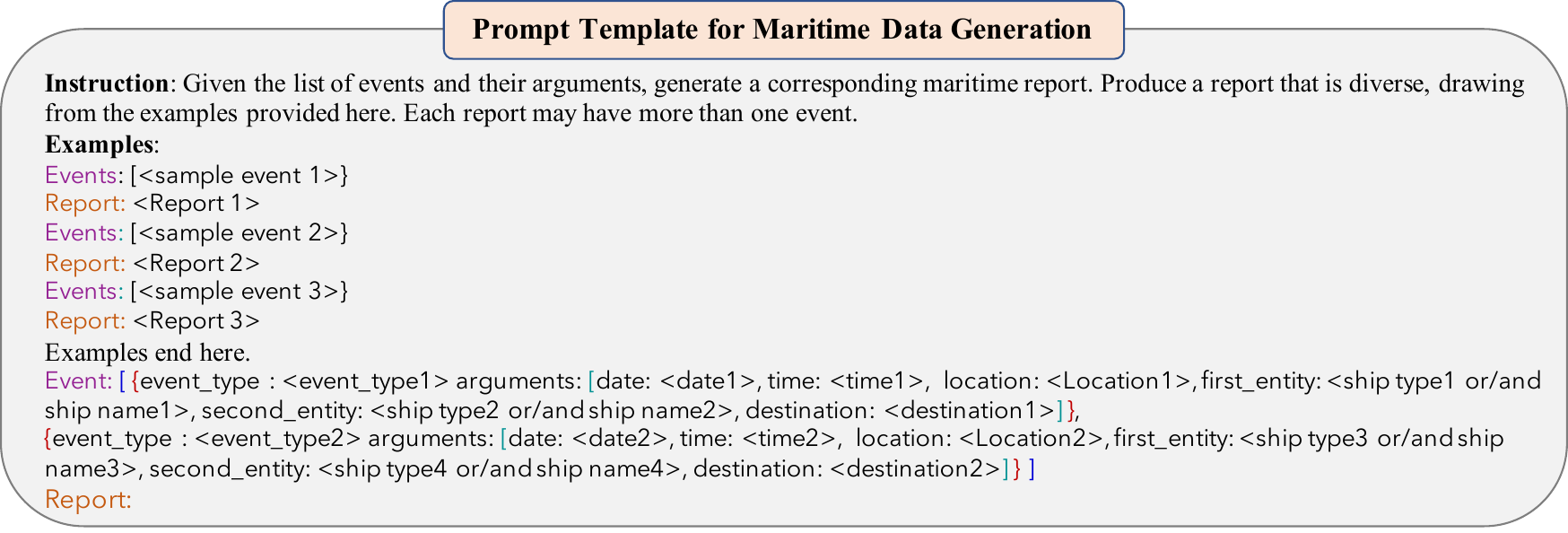}}
    \caption{Prompt template for synthesizing maritime reports.}
    \label{fig:prompt_synthsise}
\end{figure*}

\begin{figure*}[t]
    \centering
    \resizebox{0.85\textwidth}{!}{
    \includegraphics{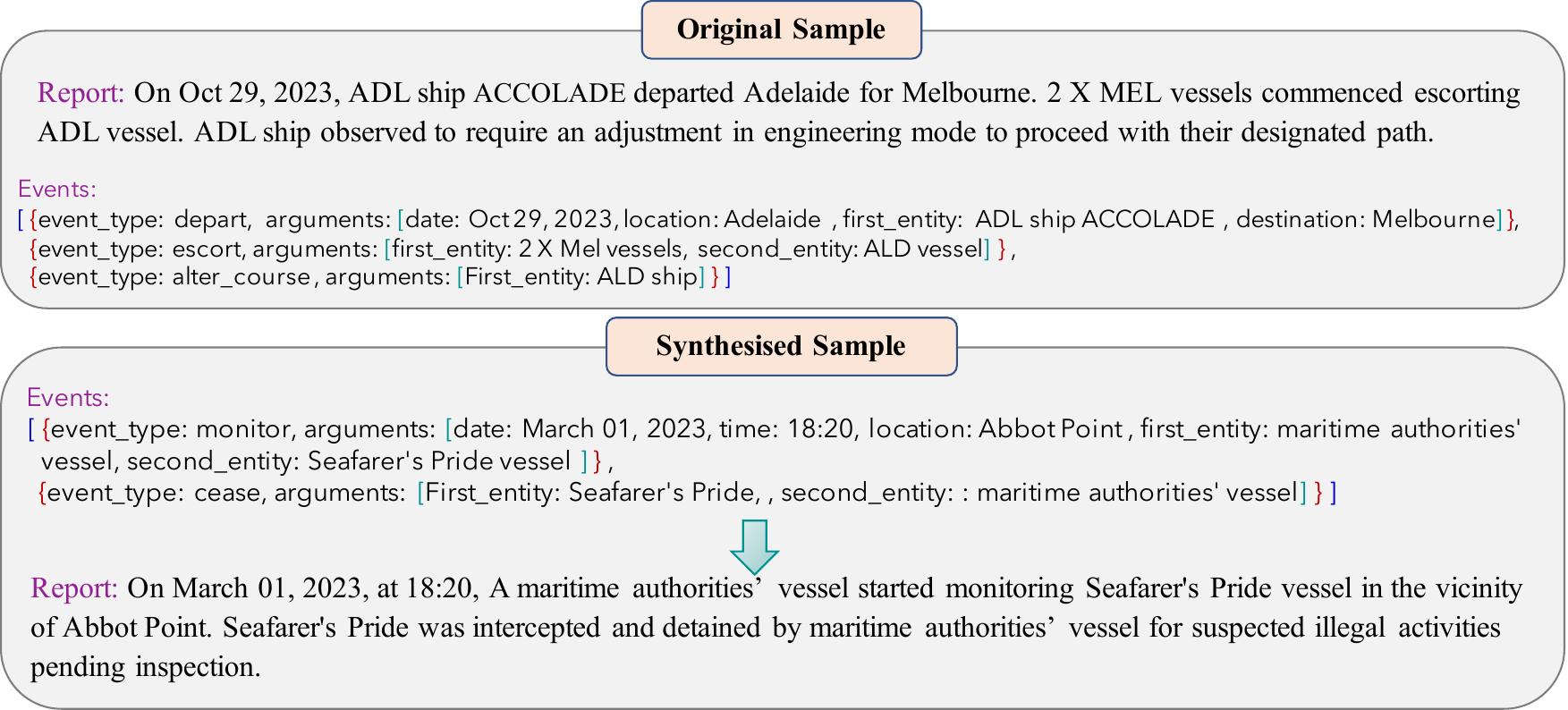}}
    \caption{Samples of original maritime reports and synthesized ones from MaritimeEvent dataset.}
    \label{fig:sample_sythesise}
\end{figure*}

Additionally, to evaluate our method in a new domain, we synthesized a new event dataset called MaritimeEvent. We collected 100 maritime-related reports containing multiple sentences and single or multiple events. We manually designed the Maritime schema, which contains 16 different event types, ${depart}$, ${arrive}$, ${monitor}$, ${kidnap}$, ${robbery}$, ${escort}$, ${block}$, ${detach}$, ${commence}$, ${transmit}$, ${cease}$, ${alter\_course}$, ${detect}$, ${pass}$, ${transit}$, and ${hail}$, and 6 argument types, ${date}$, ${time}$, ${location}$, ${ship type}$, ${ship name}$ and ${destination}$. Figure \ref{fig:sample_sythesise} illustrates two maritime samples. We use ChatGPT (gpt-3.5-turbo via OpenAI API) to synthesize approximately 10,000 diverse maritime reports. The prompt template for synthesizing a report is shown in Figure \ref{fig:prompt_synthsise}. To generate each report, we begin by randomly selecting a subset of samples from the 100 available demonstration examples as seed examples. Additionally, we choose one or a few event types, along with their corresponding arguments, from the provided lists to replace the placeholders in the prompt template. For the experiments, the temperature of ChatGPT (gpt-3.5-turbo) is set to 1.6 to synthesize maritime reports with relatively high diversity. We set the maximum number of tokens to 4000. Figure \ref{fig:sample_sythesise} illustrates an original maritime report and a synthesized one from the MaritimeEvent dataset.
The statistics of all datasets are provided in Table~\ref{tab: data}. 


\begin{table}[!htb]
\centering
\caption{\label{tab: data} Statistics of the event extraction datasets used in the paper, including the type of instances, events per instance, and the number of instances in different splits.}
\resizebox{.5\textwidth}{!}{
\begin{tabular}{lcccll}
\toprule
\textbf{Dataset} &  \textbf{Train} & \textbf{Dev} & \textbf{Test} & \textbf{Instance} & \textbf{${\#}$Events} \\ \midrule
ACE05-EN & 17,172 & 923  & 832 & One sentence & Single/Multiple \\
\textsc{WikiEvents} & 206 & 20 & 20 & Document & Multiple \\
MaritimeEvent & 9851 & 8851 & 1000 & Multi-sentence & Multiple \\\bottomrule
\end{tabular}
}

\end{table}

\textbf{Evaluation Metrics.} The results are reported using F-measure (F-1) score metrics for both event detection (Trig-C) and argument extraction (Arg-C). Trig-C indicates trigger identification and event type classification. Arg-C indicates argument identification and role classification.

\textbf{Baselines.} To validate the proposed approach through experimental comparison, we selected the following event extraction models as the baselines:
\begin{itemize}
\item Text2Event~\cite{lu2021text2event}: We compare our method with Text2Event, which is a framework that utilizes T5 models \cite{raffel2020exploring} to approach event extraction by framing it as a sequence-to-sequence generation task. In this method, all triggers, arguments, and their corresponding labels are generated as natural language words.

\item OntoGPT~\cite{ontogpt} is a tool for extracting knowledge from text by recursively querying an LLM such as GPT-3 to obtain responses using zero-shot learning. OntoGPT applies a knowledge schema on the input text and returns information conforming to the  knowledge schema. To utilize this tool for EE, we manually created the LinkML \cite{linkml} schema for the datasets for the evaluation of the OntoGPT model.

\item Fine-Grained IE\cite{gao2023benchmarking}: We compare our method with the Fine-Grained IE approach that benchmarks LLMs for Information Extraction (IE) by incorporating detailed instructions and examples for each information type. 
\end{itemize}

\begin{table*}[!htb]
\centering
\caption{\label{tab:main}Experiment results on three datasets. Trig-C indicates trigger identification and classification. Arg-C indicates argument identification and classification. The evaluation metric is F1-score. *: this model was fine-tuned on the training dataset. The LLMs were tested using direct in-context learning and were not trained or fine-tuned.}

\begin{tabular}{llcccccccc}
\hline
\multicolumn{10}{c}{\textbf{ACE205-EN}} \\ \hline
\textbf{}                                                      & \textbf{Prompt type}    & \multicolumn{2}{c}{\textbf{Zero-shot}}                                   & \multicolumn{2}{c}{\textbf{One-shot}}                                    & \multicolumn{2}{c}{\textbf{5-shot}}                                      & \multicolumn{2}{c}{\textbf{5-shot RAE}}                                  \\ \cline{2-10} 
\textbf{Model}                                                 & \textbf{Language Model} & \multicolumn{1}{l}{\textbf{Trig-C}} & \multicolumn{1}{l}{\textbf{Arg-C}} & \multicolumn{1}{l}{\textbf{Trig-C}} & \multicolumn{1}{l}{\textbf{Arg-C}} & \multicolumn{1}{l}{\textbf{Trig-C}} & \multicolumn{1}{l}{\textbf{Arg-C}} & \multicolumn{1}{l}{\textbf{Trig-C}} & \multicolumn{1}{l}{\textbf{Arg-C}} \\ \hline
Text2Event *~\cite{lu2021text2event}  & T5-large     & 71.9     & 53.8 & \-     & \-     & \-     & \-    & \-     & \-  \\ \cdashline{2-10}  
                                                               & ChatGPT                 & 43.58                               & 20.56                              & 52.38                               & 31.7                               & 66.53                               & 38.55                              & 70.43                               & 44.68                              \\
\multirow{-2}{*}{OntoGPT~\cite{ontogpt}} & GPT-4                   & \textbf{50.14}                      & 24.79                              & 60.46                               & 43.04                              & 72.92                               & 47.09                              & 77.91                               & 51.07                              \\ \cdashline{2-10} 
Fine-Grained IE\cite{gao2023benchmarking}                    & ChatGPT                 & 37.38                               & 7.39                               & 57.87                               & 14.64                              & 71.17                               & 34.40                              & \-                                 & \-                                \\ \cdashline{2-10} 
Ours$_{\text{without Decomp}}$  & ChatGPT   & 38.08    & 7.39             & 56.11      & 16.61     & 68.05      & 33.87     & 69.19  & 45.43    \\ \cdashline{2-10}
\multicolumn{1}{c}{}    & Llama2-7B               & 13.46                               & 5.31                               & 18.67                               & 9.63                               & 23.96                               & 16.61                              & 30.49                               & 18.58                              \\
\multicolumn{1}{c}{}                                           & ChatGPT                 & 44.08                               & 19.96                              & 58.33                               & 32.80                              & 71.74                               & 42.88                              & 74.55                               & 50.07                              \\
\multirow{-3}{*}{Ours}                        & GPT-4                   & 49.87                               & \textbf{25.98}                     & \textbf{63.27}                      & \textbf{45.51}                     & \textbf{75.91}                      & \textbf{51.95}                     & \textbf{81.09}                      & \textbf{58.24}                     \\ \hline \hline
\multicolumn{10}{c}{\textbf{MaritimeEvent}}  \\ \hline
\textbf{}                                                      & \textbf{Prompt type}    & \multicolumn{2}{c}{\textbf{Zero-shot}}                                   & \multicolumn{2}{c}{\textbf{One-shot}}                                    & \multicolumn{2}{c}{\textbf{5-shot}}                                      & \multicolumn{2}{c}{\textbf{5-shot RAE}}                                  \\ \cline{2-10} 
\textbf{Model}                                                 & \textbf{Language Model} & \multicolumn{1}{l}{\textbf{Trig-C}} & \multicolumn{1}{l}{\textbf{Arg-C}} & \multicolumn{1}{l}{\textbf{Trig-C}} & \multicolumn{1}{l}{\textbf{Arg-C}} & \multicolumn{1}{l}{\textbf{Trig-C}} & \multicolumn{1}{l}{\textbf{Arg-C}} & \multicolumn{1}{l}{\textbf{Trig-C}} & \multicolumn{1}{l}{\textbf{Arg-C}} \\ \hline
Text2Event *~\cite{lu2021text2event} & T5-large    & 78.52    & 59.34          & \-     & \-     & \-     & \-    & \-     & \-  \\ \cdashline{2-10}  & ChatGPT                 & 51.63    & 36.99    & 64.63   & 42.49                              & 68.89     & 46.5  & 71.25  & 50.94   \\
\multirow{-2}{*}{OntoGPT\cite{ontogpt}} & GPT-4& 55.04  & \textbf{45.05}     & 69.04    & 48.15  & 74.69     & 51.69   & 80.41    & 55.90  \\ \cdashline{2-10} 
Ours$_{\text{without Decomp}}$  & ChatGPT   & 38.08    & 7.39             & 56.11      & 16.61     & 69.56      & 33.87     & 69.99  & 49.43    \\ \cdashline{2-10}
\multicolumn{1}{c}{}  & Llama2-7B    & 18.78    & 9.56  & 25.78   & 14.56   & 31.05  & 18.92   & 40.97     & 24.87   \\
\multicolumn{1}{c}{}   & ChatGPT                 & 50.56      & 38.17    & 63.56    & 45.17                              & 68.32                               & 49.28                              & 74.39                               & 55.67                              \\
\multirow{-3}{*}{Ours}                      & GPT-4                   & \textbf{55.23}                      & 44.78                              & \textbf{70.23}                      & \textbf{50.78}                     & \textbf{75.61}                      & \textbf{55.33}                              & \textbf{84.32}                      & \textbf{60.79}                     \\ \hline \hline
\multicolumn{10}{c}{\textbf{WikiEvent}}  \\ \hline
\textbf{}                                                      & \textbf{Prompt type}    & \multicolumn{2}{c}{\textbf{Zero-shot}}                                   & \multicolumn{2}{c}{\textbf{One-shot}}                                    & \multicolumn{2}{c}{\textbf{5-shot}}                                      & \multicolumn{2}{c}{\textbf{5-shot RAE}}                                  \\ \cline{2-10} 
\textbf{Model}                                                 & \textbf{Language Model} & \multicolumn{1}{l}{\textbf{Trig-C}} & \multicolumn{1}{l}{\textbf{Arg-C}} & \multicolumn{1}{l}{\textbf{Trig-C}} & \multicolumn{1}{l}{\textbf{Arg-C}} & \multicolumn{1}{l}{\textbf{Trig-C}} & \multicolumn{1}{l}{\textbf{Arg-C}} & \multicolumn{1}{l}{\textbf{Trig-C}} & \multicolumn{1}{l}{\textbf{Arg-C}} \\ \hline
Text2Event *~\cite{lu2021text2event}  & T5-large                & 29.34                               & 17.41  & \-     & \-     & \-     & \-    & \-     & \-  \\ \cdashline{2-10}  
                                                               & ChatGPT                 & 33.67                               & 19.75                  & 46.68                   & 32.55               & 54.93  & 32.89   &55.69  & 40.9 \\
\multirow{-2}{*}{OntoGPT\cite{ontogpt}} & GPT-4                   & 41.55                               & \textbf{29.67}                     & 55.65                               & \textbf{43.78}                     & 60.99                               & 45.87                              & 61.38                               & 44.91                              \\ \cdashline{2-10} 
\multicolumn{1}{c}{}                                           & Llama2-7B               & 10.30                               & 6.42                               & 16.30                               & 9.72                               & 21.06                               & 13.58                              & 20.55                               & 14.05                              \\
\multicolumn{1}{c}{}                                           & ChatGPT                 & 39.08                               & 24.96                              & 48.08                               & 36.85                              & 56.73                               & 40.96                              & 57.89                               & 41.07                              \\
\multirow{-3}{*}{Ours}                       & GPT-4                   & \textbf{42.66}                      & 29.39                              & \textbf{56.27}                      & 40.68                              & \textbf{63.92}                      & \textbf{45.60}                              & \textbf{64.65}                      & \textbf{45.96}                    
\end{tabular}
\end{table*}

\section{Evaluation}\label{sec:eval}
This work aims to improve the automated extraction of events from text by leveraging LLMs. First, we utilize the global knowledge of LLMs through zero-shot prompts with minimal context for EE. Secondly, we evaluate LLMs in a few-shot settings in which we add a few demonstration examples to guide LLMs to extract more accurate structured events from text. Finally, we show that when combined with local knowledge through dynamic retrieval augmented examples (the most related examples to each query text), LLMs are effective in the field of EE.
Table \ref{tab:main} shows the impact of different types/number of demonstration examples within the prompt for both ED and EAE steps on three datasets. 

In zero-shot prompting, the LLM is provided with direct instruction without any additional examples. LLMs show a clear ability to learn from the context within prompts, underscoring the importance of incorporating relevant examples. In few-shot prompting we extend the number of examples to one and five canonical examples, aiming to provide additional context for the model to learn from. When applying retrieval augmented examples, both the F1 score of DE and EAE improves significantly across all LLMs, observed consistently in both ACE05-EN and MaritimeEvent datasets, characterized by a large number of instances in the training set. However, the gap remains small for the WikiEvent dataset due to a very small number of available instances (total of 206).

\section{Analysis}\label{sec:analysis}
In this section, we conduct comprehensive studies to analyze the design of our method from different perspectives.

\subsubsection{Impact of Decomposed Prompting}  \label{sec:decompose}
For a complex and challenging EE task, to enhance the performance of our approach, we design effective demonstration examples using RAE, and further, decompose the task into simpler sub-tasks via prompting. Table \ref{tab:main} presents Ours$_{\text{without Decomp}}$, indicating the performance of our proposed approach utilizing gpt-3.5-turbo without prompt decomposition across two datasets, ACE05-EN and MaritimeEvent. In this experiment, we aim to extract all event types, their triggers, and corresponding arguments in a single step. Consequently, the prompt necessitates the inclusion of all schema information, encompassing event-type definitions and their arguments. Conversely, in a two-step decomposed prompting strategy, we utilize the event type definitions in the ED prompt (Step 1) to detect the event type. In the second step, we only incorporate the schema information of the detected events in the previous step in the EAE prompt (Step 2). It is important to note that for document-level EE, a full prompt would be excessively lengthy, requiring the utilization of a costly OpenAI model (e.g., gpt-4-32k) for execution. For instance, employing decomposed prompting in a 5-shot RAE setting using ChatGPT(gpt-3.5-turbo) can improve F1 scores for ED and EAE in the ACE05-En dataset by 8.3 and 4.64 points, respectively.

\subsubsection{Impact of Embedding Methods in Retrieval Augmented Examples}
\begin{table}[]
\centering
\caption{\label{tab:embedding}Impact of embedding strategy on performance of our approach using GPT-4-turbo. ACE05-EN is used for this experiments. }
\begin{tabular}{lccc}
\textbf{Embedding   Model} & \multicolumn{1}{l}{\textbf{Embedding Vector(size)}} & \multicolumn{1}{l}{\textbf{Trig-C}} & \multicolumn{1}{l}{\textbf{Arg-C}} \\ \hline
USE~\cite{cer2018universal}               & 512                                  & 77.52                      & 56.78                     \\
RoBERTa-base~\cite{roberta}      & 768                                  & 78.87                      & 56.34                     \\
ADA-002~ \cite{ADA002}          & 1536                                 & \textbf{81.09} & \textbf{58.24}                   
\end{tabular}
\end{table}
We utilize three different state-of-the-art embedding techniques to encode the existing instances as follows:
\begin{itemize}
\item \textbf{text-embedding-ada-002 (ADA-002)}. ADA-002 \cite{ADA002} is OpenAI’s second-generation embeddings-as-a-service API endpoint model specifically adapted for text embeddings. By default, the size of the embedding vector of ADA-002 is 1536. ADA-002 is recommended by OpenAI for text similarity tasks. To retrieve the most relevant instances,  ADA-002 uses the cosine similarity between the embedding vectors of the query instance and each instance in the training dataset and returns the highest-scored instances.

\item \textbf{Universal Sentence Encoder (USE)}. USE~\cite{cer2018universal} is a widely used sentence encoding model released by Google that provides sentence-level embedding vectors implemented using a transformer. USE embeds each sentence with a 512 vector representation.

\item \textbf{RoBERTa-base}. RoBERTa-base~\cite{roberta} is a pre-trained model for text embedding. By default, it represents a sentence or paragraph by a vector with a length of 768.
\end{itemize}

Next, the encoded instance vectors are indexed using the ${IndexFlatL2}$ structure provided by Facebook AI Similarity Search (FAISS).
Table \ref{tab:embedding} illustrates the performance of our approach using gpt-4-turbo when we utilize different embedding models on the ACE05-EN dataset.
We observed that ADA-002 encodes the instances into vectors with the highest size, which is 1536. It is evident that encoding instances using the ADA-002 model achieves the highest F1 scores in both the ED and EVA sub-tasks.

\section{Conclusion }
This work addresses the challenges of structured event extraction from natural language data. Our approach leverages the capabilities of LLMs for textual understanding and enhances the accuracy of extracted events by integrating prompt enrichment techniques inspired by prompt decomposition and Retrieval Augmented Generation. Our findings reveal that utilizing LLMs enriched with schema-aware granular instructions and the retrieved augmented examples for the query substantially enhances performance. This heightened accuracy in event extraction enhances the reliability and effectiveness of automated knowledge graph construction and visualization of knowledge graphs, thereby supporting downstream reasoning and facilitating the decision-making process.

\bibliographystyle{IEEEtran}
\bibliography{IEEEabrv,IEEE_fusion}

\end{document}